\pdfoutput=1

\documentclass[11pt]{article}

\usepackage{ACL2023}

\usepackage{times}
\usepackage{latexsym}

\usepackage[T1]{fontenc}

\usepackage[utf8]{inputenc}

\usepackage{microtype}

\usepackage{inconsolata}
\usepackage{multirow}
\usepackage{makecell}
\usepackage{graphicx}
\title{ChatGPT as a Factual Inconsistency Evaluator for Text Summarization}
\author{Zheheng Luo, Qianqian Xie\thanks{\phantom{h}Corresponding author}, Sophia Ananiadou \\
Department of Computer Science, The University of Manchester \\ \{zheheng.luo, qianqian.xie, sophia.ananiadou\}@manchester.ac.uk}

\begin{document}
\maketitle
\begin{abstract}
The performance of text summarization has been greatly boosted by pre-trained language models.
A main concern of existing methods is that most generated summaries are not factually inconsistent with their source documents.
To alleviate the problem, many efforts have focused on developing effective factuality evaluation metrics based on natural language inference, question answering, and syntactic dependency et al.
However, these approaches are limited by either their high computational complexity or dependence on annotated data.
Most recently, large language models(LLMs) such as ChatGPT have shown excellent performance in not only text generation but also language comprehension.
In this paper, we particularly explore ChatGPT's ability to evaluate factual inconsistency under a zero-shot setting by examining it on both coarse-grained and fine-grained  evaluation tasks including binary entailment inference, summary ranking, and consistency rating.
Experimental results indicate that ChatGPT generally outperforms previous evaluation metrics across the three tasks, indicating its great potential for factual inconsistency evaluation. However, a closer inspection of ChatGPT's output reveals certain limitations including its preference for more lexically similar candidates, false reasoning, and inadequate understanding of instructions.


\end{abstract}

\section{Introduction}
Recently, pre-trained language models have greatly improved the performance of automatic text summarization~\cite{liu-lapata-2019-text,lewis-etal-2020-bart,zhang2020pegasus}.
However, a major concern that has limited existing state-of-the-art text summarization methods is factual inconsistency, namely, the generated summaries containing information that is not entailed by input  documents\footnote{the problem is also referred to as unfaithfulness, we use these terms interchangeably in the following }~\cite{Wojciech2020,Maynez2020}. 
To fill the gap, significant efforts have been made in developing automatic evaluation metrics for assessing the factuality of generated summaries, such as semi-supervised method FactCC~\cite{Wojciech2020}, question-answering based approach FEQA~\cite{wang-etal-2020-asking} and QuestEval~\cite{Scialom2021}, and natural language inference (NLI) based method SummaC~\citep{laban2022summac}.
Nevertheless, existing evaluation metrics either have high computational complexity which requires training on a huge amount of data or rely on multi-model combined pipelines, putting in more uncertainties during inferences. Moreover, evaluations based on these metrics exhibit limited agreement with human assessments~\cite{Pagnoni2021}.
Inspired by the ability of pre-trained language models (PLMs) on natural language understanding and generation, a few efforts have been devoted to building data and computational-efficient evaluation metrics based on PLMs like BARTScore~\cite{yuan2021bartscore}.

Most recently, large language models (LLMs), such as GPT-3~\citep{brown2020language}, InstructGPT~\citep{Ouyang2022TrainingLM}, PaLM~\citep{chowdhery2022palm}, and BLOOM~\citep{Scao2022}, have dwarfed small scale fine-tuned models in various natural language processing tasks often requiring only few-shot or zero-shot learning.
These LLMs have demonstrated exceptional performance not only in natural language understanding and generation but also in their ability to perform inference and reasoning tasks." Specifically, equipped with explicitly designed prompts, LLMs can better solve a range of various reasoning tasks in terms of arithmetics, symbolic, and logic~\citep{Kojima2022,Wei2022}. Moreover, the most recent effort ChatGPT~\cite{ChatGPT2022} in particular has been proven to obtain strong natural language inference ability, surpassing fine-tuned pre-trained language models (PLMs) on several datasets~\citep{Zhong2023}. As a result, researchers have paid closer attention to using large language models (LLMs) to evaluate generated text. 
\citet{Kocmi2023LargeLM} investigated the use of rating-based prompts in translation evaluation and achieved better accuracy compared to other metrics across three language pairs. Inspired by this work, \citet{wang2023chatgpt} extends the method into a broader natural language generation field including summarisation, where ChatGPT shows dominating alignment with human rating on four attributes including coherence, relevance, fluency, and consistency. 
However, their experiments only use a single summarisation evaluation dataset and rather focus on exploring ChatGPT to evaluate the overall quality of generated summaries. In addition, they solely framed the evaluation as a marking task and compared the results with only general text generation metrics such as ROUGE~\cite{lin-2004-rouge} and BERTScore~\cite{zhangbertscore}, which have been proven to be not effective in assessing factual consistency~\cite{Maynez2020}.
Metrics proposed specifically for assessing inconsistency such as FactCC, DAE~\citep{Goyal2020}, SummaC have not been examined, leaving a huge gap for a thorough exploration of using ChatGPT to assess the factual consistency in text summarisation.

To fill the gap, in this paper, we conduct a preliminary study of how ChatGPT can perform in both coarse-grained and fine-grained factual inconsistency evaluation from three tasks including inconsistency detection as entailment inference (EI), consistency comparison as summary ranking, and quantitative judgement as consistency rating. We design different prompts on both zero-shot and zero-shot chain-of-thought (CoT)~\cite{Kojima2022} to explore the factuality assessment ability of ChatGPT. We conduct experiments on the benchmark of the EI-based inconsistency detection task including six large standardized datasets, and existing datasets on the other two tasks, and compare the results with SOTA evaluation methods. From experimental results and analysis, we have the following findings:
\begin{enumerate}
    \item  ChatGPT shows great potential for evaluation factuality of text summarisation under the zero-shot setting and outperforms previous SOTA evaluation methods on most datasets across three tested tasks.
    
     \item  Though showing remarkable performance measured by numeric metrics, ChatGPT is found to have the preference to predict a document and a claim is consistent when the lexical similarity is high without considering the semantic entailment between them. Moreover, evidence of ChatGPT conducting false inferences has been observed, revealing the limitation of ChatGPT's language reasoning ability.
    
    \item Despite effectively instructing ChatGPT to detect inconsistency, the tested prompts are not able to keep the output constantly sticking to the given requirements, indicating the insufficient prompting of ChatGPT.
\end{enumerate}
To the best of our knowledge, we are the first to systematically explore ChatGPT's ability in evaluating factual consistency for text summarization. Overall, our results show a comparable if not better performance of ChatGPT than SOTA evaluation metrics, but concerns remain on lexical biases, false reasoning, and inadequate alignment which are expected to be addressed to improve its reliability.

\section{Related Work}
\subsection{Factuality Evaluation in Text Summarization}
Existing factuality evaluation metrics generally can be classified into unsupervised and semi-supervised methods.
Unsupervised evaluation metrics generally include information extraction (IE) based methods, natural language inference (NLI) based methods, and question answering (QA) based methods.
\citet{goodrich2019assessing} proposed the model-based factuality evaluation metric to calculate the overlap of relation tuples (subject, relation, object) that are extracted from generated summaries and the ground truth by the information extraction (IE) model.
\citet{nan-etal-2021-entity} proposed the new evaluation metric assessing the entity-level factuality consistency of generated summaries.

Besides the IE-based methods, natural language inference (NLI) is also explored for factuality evaluation by assessing whether the generated summary is entailed by the input document.
\citet{falke-etal-2019-ranking} found the factuality evaluation methods trained on the NLI datasets have a poor ability to the assessment of text summarization.
\citet{mishra-etal-2021-looking} further found that the poor performance of the evaluation methods training with the NLI datasets is caused by the short length of premises in NLI datasets.
Most recently, \citet{laban2022summac} revisited the use of NLI in inconsistency detection by calculating the factuality score based on sentence pairs, and proposed the novel benchmark SUMMAC (Summary Consistency) with six datasets.
SUMMAC is used in our experiments.

Moreover, there is also question answering-based metrics such as FEQA~\cite{durmus-etal-2020-feqa}, QAGS~\cite{wang-etal-2020-asking}, and QuestEval~\cite{Scialom2021}, by assessing the alignment of the generated answer based on the generated summary and the source, with the given question.
Different from unsupervised NLI-based methods, the semi-supervised methods further utilize the synthetic data from text summarization for weakly supervised learning, such as FactCC~\cite{Wojciech2020}.
However, these methods are usually computationally expensive or rely on annotated data~\cite{huang2021factual}.
Inspired by the effectiveness of PLMs, there are efforts on developing factuality evaluation metrics based on the likelihoods of PLMs, that are computation and data-efficient such as BARTScore~\cite{yuan2021bartscore} and T5Score~\cite{qin2022t5score}.

\subsection{ChatGPT for Natural Language Processing}
Most recently, many efforts have explored the zero-shot ability of ChatGPT on various natural language processing tasks~\cite{jiao2023chatgpt,Zhong2023,qin2023chatgpt,bang2023multitask,yang2023evaluations}.
ChatGPT has been proven to exhibit good performance on machine translation~\cite{jiao2023chatgpt}.
On the GLUE benchmark, \citet{Zhong2023} has found ChatGPT shows significantly better performance on inference tasks, has comparable performance on sentiment analysis and question-answering tasks, and has poor performance on paraphrase and similarity tasks when compared with 4 representative BERT-based fin-tuning methods.
\citet{qin2023chatgpt} further shows ChatGPT has superior performance on reasoning-required tasks including dialogue tasks, natural language inference tasks, and question-answering tasks than GPT-3.5, and has worse performance on the summarization task than GPT-3.5.
ChatGPT and GPT-3.5 have comparable performance on sentiment analysis.
\citet{bang2023multitask} shown ChatGPT outperforms SOTA zero-shot methods in 9/13 NLP datasets and has poor performance on low-resource languages such as Marathi, Sundanese, and Buginese.
\citet{yang2023exploring} and \citet{wang2023cross} explored the query and aspect-based text summarization and cross-lingual summarization with ChatGPT, where it shows comparable performance with the fine-tuning-based methods.
\cite{Soni2023ComparingAS} conducted a human evaluation and found reviewers struggle to distinguish hand-written summaries against generated ones from ChatGPT.
\citet{wang2023chatgpt} examined the ability of ChatGPT on evaluating natural language generation (NLG) tasks such as summarization, story generation, and data-to-text tasks.
ChatGPT shows great potential as the NLG metric, whose evaluation results have a high correlation with human judgment.
However, they only utilized one summarisation dataset and focused on exploring ChatGPT to evaluate the relevance by comparing it with non-factuality evaluation metrics such as ROUGE and BERTScore, leaving a huge gap for a thorough exploration of the ability of ChatGPT to assess the factual consistency in text summarisation.

\section{ChatGPT as a Factual Inconsistency Evaluator}
In this section, we introduce the details of three different tasks for detecting inconsistency with ChatGPT including the prompt designing, evaluation setting, tested datasets, and baseline models.

\subsection{Entailment Inference}
\label{task intro: nli}
\noindent\textbf{Evaluation Setting.} Inconsistency evaluation of the generated summary can be cast as a binary natural language inference classification, in which the evaluation model is solely required to assess if the summary is consistent with the source document rather than rating the consistent levels~\cite{laban2022summac}.
Under this framework, two parameters are needed for the prompts: source document and summary.
We provide ChatGPT with the question including the source document and the corresponding generated summary and ask it to answer yes or no to infer the consistency between the source document and the corresponding generated summary, and then we collect the decisions from the outputs and aggregate the results.

\noindent\textbf{Prompts.} We experiment with two different zero-shot prompts in the NLI setting. The first one is based on \textit{direct assessment} by directly asking ChatGPT to answer yes or no given the question. Another is based on \textit{zero-shot Chain-of-Thought} inspired by previous work~\citep{Kojima2022} of adding "let's think step by step" in prompt to encourage LLMs unfolding a chain-of-thought style reasoning process, which has been proved to be effective on several reasoning tasks. We follow the approach to create the second prompt.
The zero-shot template is shown below:
\begin{quotation}
Decide if the following summary is consistent with the corresponding article. Note that consistency means all information in the summary is supported by the article. 
    
    Article: \textcolor{blue}{[Article]}
    
    Summary: \textcolor{blue}{[Summary]}
    
    Answer (yes or no):
\end{quotation}
The zero-shot CoT template is:
\begin{quotation}
Decide if the following summary is consistent with the corresponding article. Note that consistency means all information in the summary is supported by the article. 
    
Article: \textcolor{blue}{[Article]}
    
Summary: \textcolor{blue}{[Summary]}
    
Explain your reasoning step by step then answer (yes or no) the question:
\end{quotation}

When processing the responses, we only consider solid judgment like "the summary is consistent with the article" as consistency, claims such as "partially consistent" or 'mostly consistent' are all deemed as inconsistent.
We also tried to use few-shot prompts. However, we found the performance unstable when changing the label, order, and amount of examples, so we decide to leave it for further exploration.

\noindent\textbf{Datasets} We evaluate ChatGPT's performance on the SUMMAC benchmark~\citep{laban2022summac} which includes six largest summary inconsistency detection datasets FactCC~\cite{Wojciech2020}, CoGenSumm~\cite{falke-etal-2019-ranking}, XSumFaith~\cite{Maynez2020}, SummEval~\citep{Fabbri2021}, FRANK~\cite{Pagnoni2021}, and Polytope~\citep{huang-etal-2020-achieved}. Notably, not all the datasets in the SUMMAC benchmark are built for binary consistency classification. For example, in SummEval~\citep{Fabbri2021}, generated summaries are marked on consistency over a range from 1-5 points. SUMMAC standardizes the six datasets into a binary classification format where each instance contains a triplet of (document, summary, label). The label is either consistent or inconsistent. Moreover, they manually created validation and test split for datasets where a such split is not conducted and computed the inter-annotator agreement for data with multiple annotators. The statistics of the benchmark are shown in Table \ref{tab: summac}.

\noindent\textbf{Baseline Models.}
We compare ChatGPT's performance with the following methods:
\begin{itemize}
    \item \textbf{NER Overlap} uses the named entity
recognition (NER) model
to detect inconsistency by examining if an entity in the summary is
in the document~\citep{laban-etal-2021-keep}. The tested model considers only a subset of entity types such as PERSON,
LOCATION, ORGANIZATION, etc.

\item \textbf{MNLI-doc} fine-tunes a Roberta model~\citep{Liu2019RoBERTaAR}
on the MNLI dataset~\citep{williams-etal-2018-broad} and labels the document-summary pair by the predicted probability of entailment.

\item \textbf{FactCC}~\citep{Wojciech2020} is a Roberta model fine-tuned on data synthesized by corrupting sentences in the original documents as inconsistent candidates.

\item \textbf{DAE}~\citep{Goyal2020} is a parsing-based model evaluating inconsistency by examining the entailment of individual dependency arcs.

\item \textbf{FEQA}~\citep{Durmus2020} first generates question-answer pairs from candidate summaries, then compare the answers extracted from the source documents by asking the same questions. Then the answer sets are compared to determine the consistency.

\item \textbf{QuestEval}~\citep{Scialom2021} extends the methods above by adding an information recall score to a QA-based metric.

\item \textbf{SummaC}~\citep{laban2022summac} builds an NLI matrix by splitting the document and summary into sentence sets, then predicts a score for each sentence pair in the matrix. SummaC zero-shot ($Summac_{zs}$) first obtain the maximum along the columns then average over to get a final consistency score. SummaC convolution ($SummaC_{Conv}$) instead trains a convolution layer to predict a score for each column and then uses the mean output as the summary-level score.
\end{itemize}

Detailed implementations of the above models used to compare can be found in~\cite{laban2022summac}. For scoring models, the threshold is selected using the validation set and allowed to vary over different datasets.

\noindent\textbf{Metric.} Due to the unbalanced distribution of positive and negative samples in the testing sets, we choose balanced accuracy~\citep{brodersen2010balanced} as the main metric since it is more sensitive to predictions difference for data of smaller proportions. Balanced accuracy
is defined as the following:
\begin{equation}
bACC = \frac{1}{2}* (\frac{TP}{TP+FN} + \frac{TN}{TN+FP})
\end{equation}
The first term in the equation is sensitivity, which represents the recall of true positives while the next one is specificity standing for the recall of true negatives. We specifically counted the two sub-metrics to analyze ChatGPT's behavior.
\begin{table}[!tb]
\small
\centering
\begin{tabular}{p{0.25\linewidth}p{0.1\linewidth}p{0.1\linewidth}p{0.15\linewidth}p{0.1\linewidth}}
\hline
\textbf{Dataset} &  
\textbf{Valid. size} &\textbf{Test size}&  \textbf{\%Positive}&\textbf{Source}\\
\hline
CoGenSumm &1281&  400& 49.8 &C\\
XSumFaith &1250&1250 &10.2 &X\\
Polytope &634& 634 & 6.6&C\\
FactCC &931& 503 &85.0&C \\
SummEval &850&850 &90.6 &C \\
FRANK& 671&1575& 33.2 &C+X\\

\hline
\end{tabular}
\caption{\label{tab: summac}
Statistics of datasets in SUMMAC Benchmark.
}
\end{table}

\subsection{Summary Ranking}

\noindent\textbf{Evaluation Setting} Except binary NLI, a model's awareness of factual inconsistency can also be tested on how whether it can rank a consistent summary over an inconsistent one. In this section, we introduce another evaluation task~\textit{Summary Ranking} which is introduced in~\citet{falke-etal-2019-ranking}
and has been tested in other previous work. Specifically, the model will be asked to choose the consistent one over two candidate summaries (one is faithful, the other one is not) given the source document.

\noindent\textbf{Prompts}
We use a zero-shot prompt which directly asks ChatGPT to answer which sentence out of the two candidates is more consistent with the given article sentence.
\begin{quotation}
    Decide which of the following summary is more consistent with the article sentence. Note that consistency means all information in the summary is supported by the article.
    
    Article Sentence: \textcolor{blue}{[article]} 
    
    Summary A: \textcolor{blue}{[correct summary]}
    
    Summary B: \textcolor{blue}{[incorrect summary] }
    
    Answer (A or B):
\end{quotation}

\noindent\textbf{Dataset}
Here we use the dataset built by~\citet{falke-etal-2019-ranking} which contains
373 samples, each containing an input source
document from CNN/DM~\citep{Nallapati2016} and two summary sentences covering the same content. 
One of
the summary sentences are consistent with the article while
the other is inconsistent. 

\noindent\textbf{Baseline Models}
We compare other evaluation models that reported their performance on this dataset including the aforementioned FactCC~\citep{Wojciech2020}, MNLI-doc, DAE~\citep{Goyal2020} and a human judgement from~\cite{falke-etal-2019-ranking}.

\noindent\textbf{Metric} We report the accuracy of models successfully choosing consistent summary over inconsistent one. Specifically, when collecting responses from ChatGPT, we only deem claims that confirm the correct sentence is consistent as correct. Outputs alleging both candidate sentences are consistent or inconsistent are rendered as failures.

\subsection{Consistency Rating}
\noindent\textbf{Evaluation Setting.} Recently, several studies have found when given accordingly request prompts, LLMs are able to mark the quality of generated text from different aspects~\citep{Kocmi2023, Fu2023GPTScoreEA, Wang2023IsCA}. These scores show high correlations with human assessment, suggesting the potential of ChatGPT in predicting fine-grained consistency levels for summarisation. Moreover, in the experiments of the NLI task in Section \ref{task intro: nli}, we found that part of the output judgments is "partially consistent" or "mostly consistent", indicating ChatGPT's awareness of different inconsistency degrees.
Therefore, we apply the consistency rating task on ChatGPT by asking it to mark the consistency of a summary with the reference to its source document on a scale from 1-10 points, where 1 point stands for total inconsistency, and 10 represents full consistency.

\noindent\textbf{Prompts.} Following~\citet{Kocmi2023}'s approach, we design a prompt that requests ChatGPT to evaluate the consistency of a candidate summary w.r.t the source article in a [1-10] scale:
\begin{quotation}
Score the following summary given the corresponding article with respect to consistency from 1 to 10. Note that consistency measures how much information included in the summary is present in the source article. 10 points indicate the summary contains only statements that are entailed by the source document.

\textcolor{blue}{[Summary]: }

\textcolor{blue}{[Source Article]:}

Marks:

\end{quotation}
The definition of consistency is added for the model to better understand the aspect it is to rate.

\noindent\textbf{Datasets.} The original versions of SummEval and FRANK datasets are used on this task given there are detailed consistency scores in their annotations. In SummEval, 1600 summaries were labeled using a 5-point Likert scale along four categories: coherence, consistency, fluency, and relevance by 3 expert annotators. We average the points in the consistency aspect as the final score. FRANK has a binary consistency score for each sentence in a summary labeled by annotators, then aggregates a summary-level score from 0-1, resulting in 2250 marked summaries in total.

\noindent\textbf{Baseline Models}
We compare other evaluation models that reported their performance on this dataset including the aforementioned FactCC, FEQA, DAE and QAGS~\citep{wang-etal-2020-asking}, which is a QA-based faithfulness evaluation method.

\noindent\textbf{Metrics} To evaluate to what extent the examined models
align with human judgment. Two widely-used
correlation measures are adopted: (1) Spearman
correlation~\citep{zar2005spearman} assesses the monotonic relationships between two variables; (2) Pearman
correlation~\citep{mukaka2012guide} measures the linear relationships between two sets of data; (3) Kendall’s
Tau~\cite{kendall1938new}evaluates the ordinal association between two measured quantities.
\begin{table*}[!tb]
\centering
\begin{tabular}
{p{0.15\linewidth}m{0.1\linewidth}m{0.1\linewidth}m{0.08\linewidth}m{0.08\linewidth}m{0.1\linewidth}m{0.08\linewidth}}\\
\hline
 \multirow{2}{*}{\textbf{Methods}}&  
\multicolumn{5}{c}{\textbf{SUMMAC Benchmark Datasets}}\\
\cline{2-7}
 \textbf{}&  
\textbf{CoGenSum} &\textbf{XsumFaith}&  \textbf{Polytope}&  \textbf{FactCC}& \textbf{SummEval}&  \textbf{FRANK}\\
\hline
NER Overlap&  53.0 &63.3 &52.0 &55.0& 56.8& 60.9\\
MNLI-doc &57.6& 57.5 &61.0 &61.3 &66.6 &63.6\\
FactCC-CLS& 63.1& 57.6& 61.0 &75.9& 60.1 &59.4\\
DAE& 63.4 &50.8& 62.8& 75.9& 70.3 &61.7 \\
FEQA& 61.0 &56.0 &57.8 &53.6 &53.8 &69.9 \\
QuestEval& 62.6 &62.1 &\textbf{70.3}& 66.6& 72.5 &82.1\\
SummaC\textsubscript{ZS}&70.4 &58.4 &62.0& 83.8 &78.7& 79.0\\
SummaC\textsubscript{Conv} &64.7& \textbf{66.4} &62.7 &\textbf{89.5} &81.7 &81.6\\
ChatGPT\textsubscript{ZS} &63.3	&64.7	&56.9	&74.7&	76.5&	80.9\\
ChatGPT\textsubscript{ZS-COT}& \textbf{74.3}&	63.1	&61.4	&79.5	&\textbf{83.3}&	\textbf{82.6}\\
\hline
\end{tabular}
\caption{\label{tab: summac results}
Balanced accuracy results of inconsistency detect models on the test set of SummaC. Results of baselines are referenced from the paper~\cite{laban2022summac}.
}
\end{table*}
\section{Experiment}
We conduct our experiments using the API of ChatGPT (\textit{gpt-3.5-turbo-0301}) which is trained based on InstructGPT~\citep{Ouyang2022TrainingLM} with reinforce learning from human feedback (RLHF).
To avoid the effects of historical dialogues, we sent each request individually to obtain the response.
\subsection{Entailment Inference}
\label{sec entailment}
The full results of the entailment inference task are shown
in Table \ref{tab: summac results}. Overall, ChatGPT is able to achieve comparable performance or even better performance compared to the previous state-of-the-art evaluation models without training on relevant tasks, demonstrating the potential of ChatGPT-like LLMs on detecting inconsistency between two pieces of text in a zero-shot setting.
Specifically, ChatGPT with zero-shot CoT prompt produces the best results and outperforms the previous SOTA method SummaC\textsubscript{ZS} by 3.9\%, 1.6\% and 1.0\% on CoGenSum, SummEval, and FRANK datasets correspondingly.
It remains comparable to the best models on the rest three datasets including XsumFaith (63.1\% compared to SummaC\textsubscript{Conv} with 66.4\%), Polytope (61.4\% compared to QuestEval with 70.3\%), FactCC (79.5\% compared to SummaC\textsubscript{Conv} with 89.5\%). 
In almost all datasets, the ChatGPT\textsubscript{ZS-COT} which guides the ChatGPT with the chain-of-thought prompt has significantly better performance than ChatGPT\textsubscript{ZS}, In detail, ChatGPT\textsubscript{ZS-COT} outperforms ChatGPT\textsubscript{ZS} by 11.0\%, 4.5\%, 4.8\%, 6.8\% and 1.7\% on the CoGenSum, Polytope, FactCC, SummEval, and FRANK datasets correspondingly.
It shows great potential to better explore the factuality evaluation ability of ChatGPT by prompt engineering in the future.
\begin{figure}
    \includegraphics[width=0.5\textwidth]{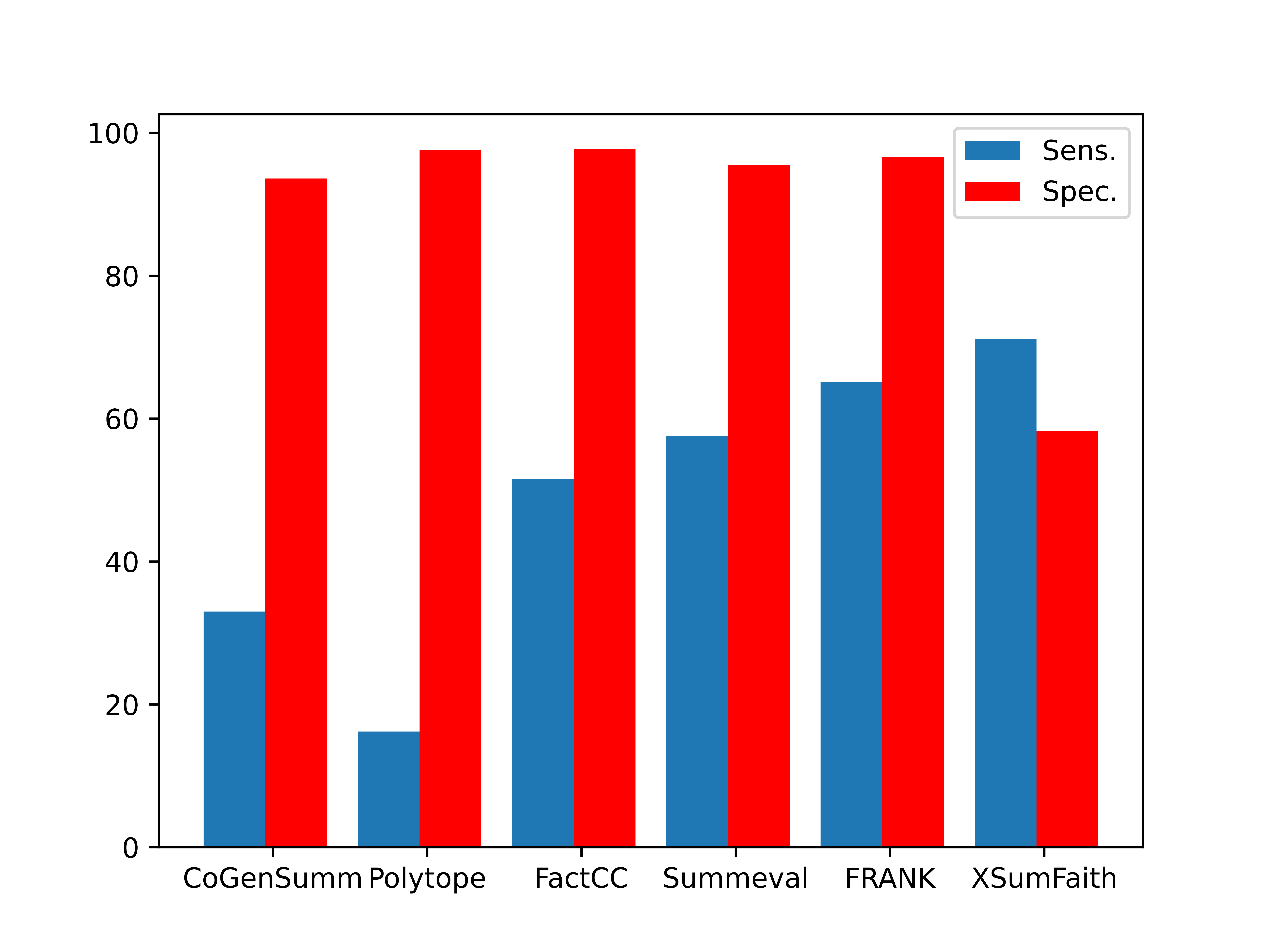}
    \caption{The results of sensitivity and specificity of ChatGPT\textsubscript{ZS-COT}.}
    \label{fig: SenVSSpec}
\end{figure}

To further investigate ChatGPT's performance in consistent and inconsistent instances, we break the balanced accuracy results of ChatGPT\textsubscript{ZS-COT} into sensitivity (positive recall) and specificity (negative recall), the comparison is in Fig \ref{fig: SenVSSpec}. In five out of the total six datasets, ChatGPT can successfully retrieve more than 95\% consistent summaries (high negative recall namely specificity), while performing rather poorly on identifying all the inconsistent ones (low positive recall namely sensitivity). Based on this observation, we assume that during inference, ChatGPT might still rely more on semantic similarity to make its decision on consistency detection since most of the candidate summaries are lexically close to sentences in the source articles, causing its vulnerability in finding these trivial modifications in inconsistent summaries that changes the meaning of the source document. This could be further demonstrated in ChatGPT's reverse performance on the two types of candidate summaries in the XSumFaith dataset which contains summaries generated by models trained on the XSum dataset in Table \ref{tab: summac results}. Previous works~\citep{durmus-etal-2020-feqa} have shown that the generated summaries are highly affected by training data and models trained on CNN/DM produce nearly extractive summaries while the same model trained on XSum will give significantly more abstractive ones. Abstrativeness brings the decline of lexical similarity between the candidate summary and the source document which might be the main reason why in XSumFaith, ChatGPT tends to predict more cases as inconsistent.
\subsection{Summary Ranking}
\begin{figure}[htb!]
    \includegraphics[scale=0.38]{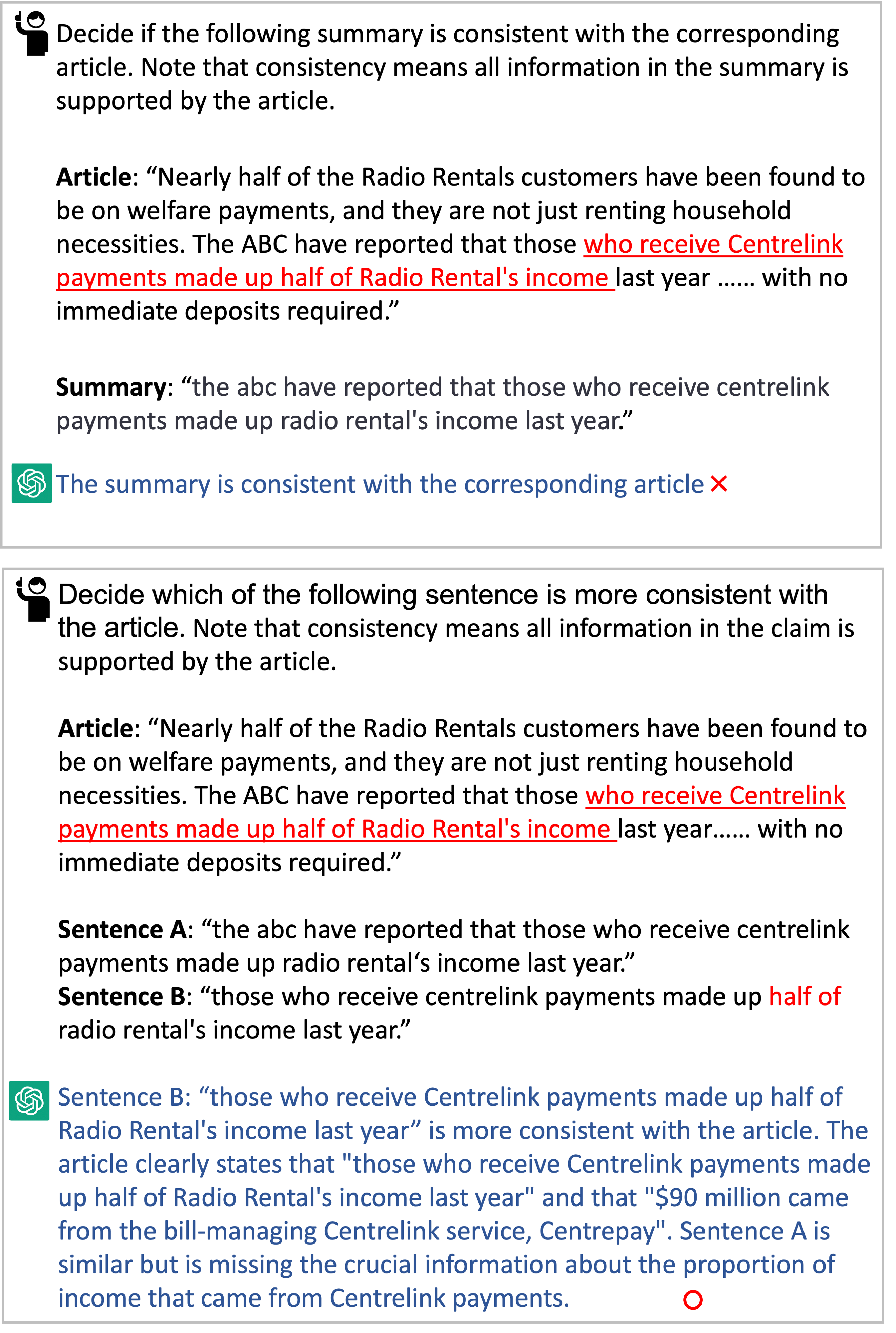}
    \caption{ChatGPT's actions when given the same source document and an inconsistent summary but with and without a consistent one. The red underlined text in the article is content highly related to the candidate summaries.}
    \label{fig: fail EL got SR}
\end{figure}

\begin{table}
\centering
\begin{tabular}{p{0.5\linewidth}m{0.3\linewidth}}
\hline
\textbf{Model} &  
\textbf{Ranking Acc.}\\
\hline
FactCC &70.0\\
MNLI-doc &78.3\\
Rule-based dependency &74.8\\
DAE &83.6\\
Human &83.9\\
ChatGPT & \textbf{85.2}\\
\hline
\end{tabular}
\caption{\label{tab: summary ranking}
Performance of models on the
summary ranking task. Results of baselines are reported in~\citet{Goyal2020}.
}
\end{table}
The results of the summary ranking task are shown in Table \ref{tab: summary ranking}. It shows that ChatGPT without any in-context learning can outperform not only existing methods but also a human assessment reported in~\citet{Falke2019}. 
Notably, the ranking dataset is sampled from the output of models trained on CNN/DM. Therefore, the candidate summaries are mostly identical to some sentences in the source document, the inconsistent ones tend to contain minor adjustments corrupting the meaning like deleting modifiers like "half of" as shown in Figure \ref{fig: fail EL got SR}. Though we conclude from Section \ref{sec entailment} that ChatGPT relies heavily on lexical similarity to decide the consistency degree of sentences, in this summary ranking task, we see that ChatGPT can detect the trivial semantic differences even when given two highly similar candidates and pick out the consistent one.
For example, in the second case of Figure \ref{fig: fail EL got SR}, ChatGPT can correctly assess that sentence B is more consistent with the input article, given the highly lexical similarity between sentence B and sentence A.
In our manual inspection, we found that ChatGPT is able to point out the inconsistency in some cases where it failed in the entailment inference when ranking them compared to their consistent counterparts.
As shown in the first case of Figure \ref{fig: fail EL got SR}, ChatGPT failed in detecting the inconsistency of the summary with the input article in the entailment inference task.
While it can correctly pick out the more consistent one when given two summaries with highly lexical similarity in the summary ranking task, as shown in the second case of Figure \ref{fig: fail EL got SR}.
This indicates the importance of prompt engineering with useful contexts in better triggering ChatGPT's capability.

\begin{table*}
\centering
\small
\begin{tabular}{m{0.07\linewidth}m{0.07\linewidth}m{0.07\linewidth}|m{0.07\linewidth}m{0.07\linewidth}|m{0.07\linewidth}m{0.07\linewidth}|m{0.07\linewidth}m{0.07\linewidth}}\\
\hline
&\multicolumn{2}{c}{\textbf{FRANK}}&\multicolumn{2}{c}{\textbf{FRANK(CNN/DM)}}&\multicolumn{2}{c}{\textbf{FRANK(XSum)}}&\multicolumn{2}{c}{\textbf{SummEval}}\\\hline
\multirow{2}{*}{Metrics}& Pear. &Spear.& Pear. &Spear.& Pear. &Spear.&Pear. &Spear.\\
&$\rho$ &$r$&$\rho$ &$r$ &$\rho$ &$r$ &$\rho$ &$r$\\\hline
FEQA &0.00& 0.01&-0.01& -0.01 &0.02&0.07 &-&-\\
QAGS&0.06&0.08&0.13&0.09&-0.02&0.01&-&-\\
DAE&0.16&0.14&0.25&0.24&0.04&\textbf{0.28}&0.20&0.27\\
FactCC&0.20&0.30&0.36&0.33&0.07&0.25&0.32&0.34\\
ChatGPT&\textbf{0.70}&\textbf{0.69}&\textbf{0.50}&\textbf{0.46}&\textbf{0.34}&0.27&\textbf{0.49}&\textbf{0.35}\\\hline
    \end{tabular}
    \caption{
    \label{tab: correlation}
    Pearson correlation, and spearman rank correlation coefficients 
    between human judgements and evaluation scores of different methods.}
\end{table*}

\subsection{Consistency Rating}
We further show the performance of all methods on the consistency rating task in Table \ref{tab: correlation}, where we compare the correlations of their rating results with human judgement. Still, without in-context training, ChatGPT outperforms other consistency metrics by aligning closer to human assessments. Especially in the whole FRANK dataset, ChatGPT leads other metrics by a large margin, emphasising its superior ability in measuring the consistency degree than the baseline models.

In particular, when splitting the FRANK dataset into summaries from CNN/DM and XSum, the correlations of ChatGPT show a considerable decline from CNN/DM to XSum, which matches our analysis in the previous two parts. The difference might come from the abstractiveness of summaries generated from models trained on XSum, so their lower lexical similarity with the source document affects the model's judgement of consistency, leading to the worse performance in the FRANK XSum dataset. However, though the abstractiveness of XSum summaries lowers the correlations generally, ChatGPT's pearson's correlation is still much higher than the single-digit results of the baselines, suggesting its better language understanding and inference ability.
\begin{figure}
    \centering
    \includegraphics[scale=0.36]{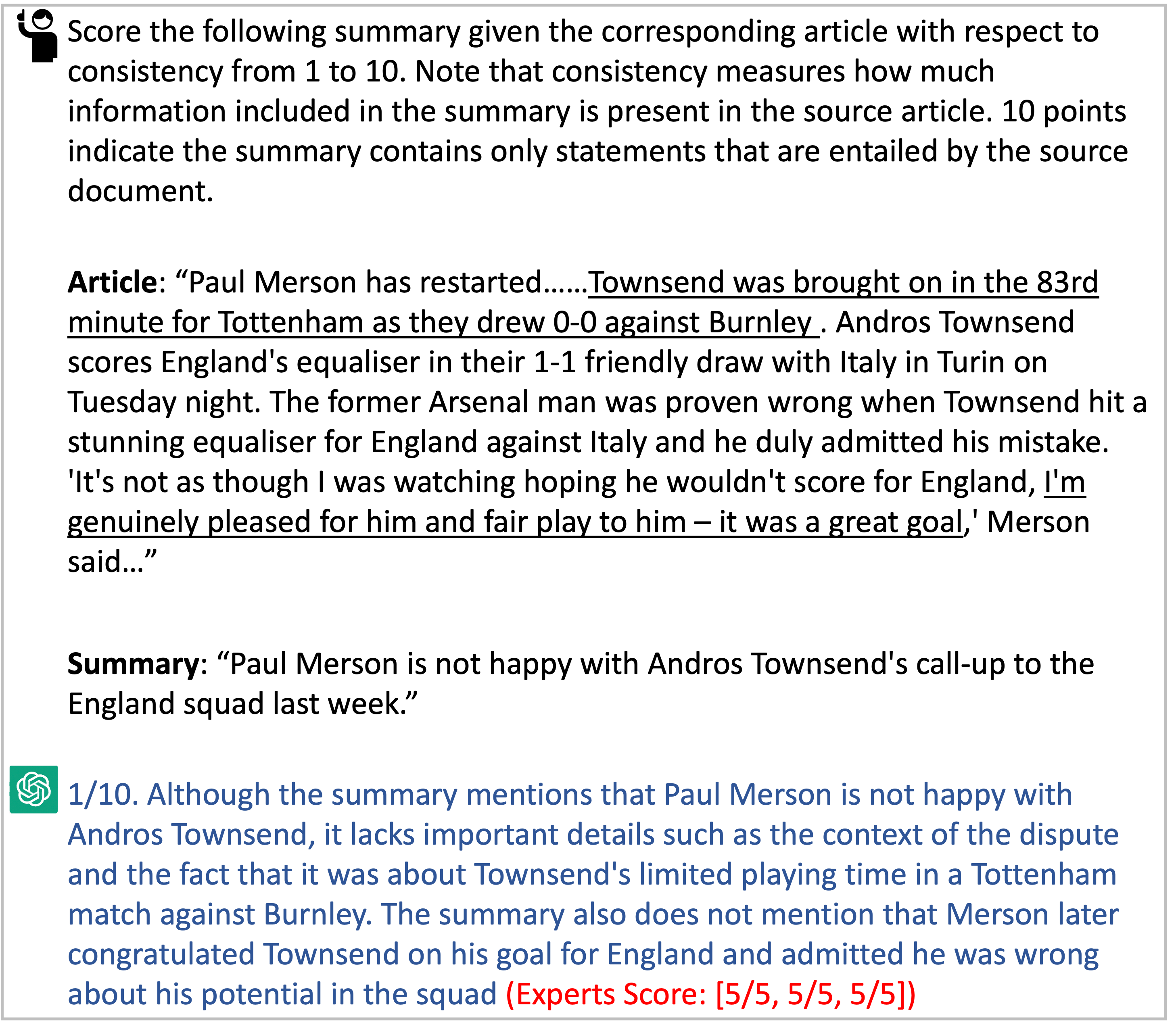}
    \caption{An example of ChatGPT fail to stick to the given definition of consistency.}
    \label{fig:rating failure}
\end{figure}
\subsection{Error Analysis}
In this part, we show some example cases of ChatGPT in the three tasks to showcase its limitations and attempt to provide a hint to understand ChatGPT's behavior in the aforementioned tasks. 

In Figure \ref{fig: fail EL got SR}, we show an example from the CoGenSumm dataset where ChatGPT failed in the entailment inference task. The model neglects the disappearance of "half of " in the candidate summary which significantly changes the meaning and decides the summary is consistent with the article.
However, when putting the same summary and article into the summary ranking task combined with a consistent claim, ChatGPT successfully picks the consistent one and gives the right reasoning of why "Summary A" is inconsistent.
The first case of Figure \ref{fig: fail EL got SR} supports our assumption of ChatGPT counting on lexical similarity to determine consistency as the high lexical overlap between inconsistent summary and the red-underlined part in the article cheats ChatGPT. Nevertheless, when another summary is both lexically and semantically closer to the article, ChatGPT detects the difference and manages to answer correctly in the second case of Figure \ref{fig: fail EL got SR}.

\begin{figure}
    \centering
    \includegraphics[scale=0.4]{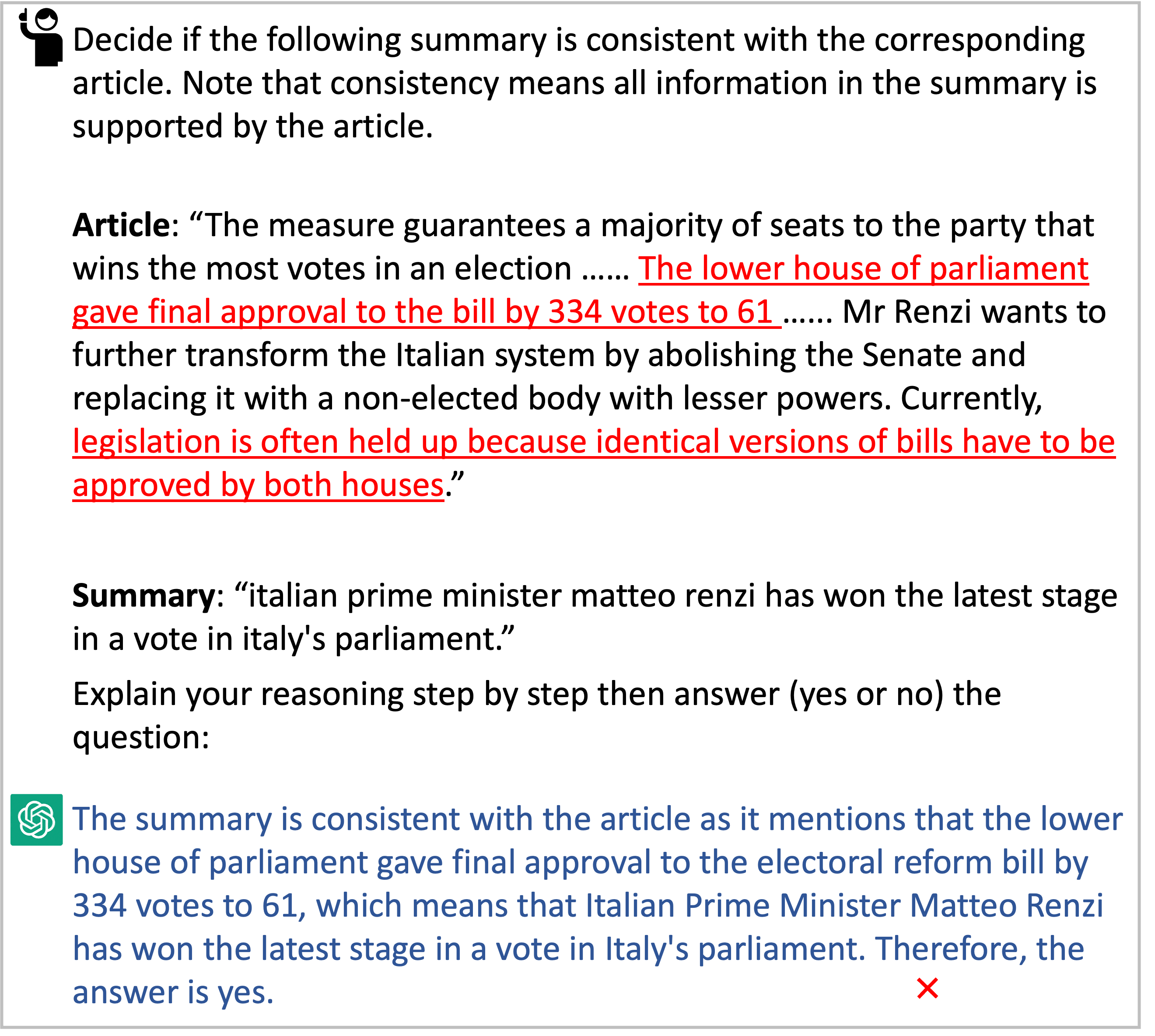}
    \caption{An example of ChatGPT conducts false reasoning.}
    \label{fig: wrong reasoning}
\end{figure}

With further investigation of failure cases, we found ChatGPT makes false inferences as shown in Figure \ref{fig: wrong reasoning}. The summary claims that "prime minister matteo renzi" won the vote while the red underlined part in the article clearly says the bill has passed the lower house but is held up to be approved by both houses. However, ChatGPT determines this summary is consistent and tries to justify it by using "the bill is approved by the lower house" as evidence. This example, combined with the upper case in the first example, demonstrates that ChatGPT still has a limitation on understanding and inferencing of natural language. Furthermore, a CoT-style prompt is applied in this example to encourage the model to generate a reasoning process to assist its judgment. But ChatGPT directly produces the conclusion first and then unfolds its inference progress afterwards. According to the autoregressive training nature of GPT, the explanation is then conditioned on the "consistent" conclusion and thus cannot guide the decision while following the judgment. In our manual inspection, answers with the conclusion at first are not rare, suggesting zero-shot CoT-style prompts might not be the optimal instruction for ChatGPT to conduct a language inference task with reasoning progress. We suppose fined-engineered few-shot prompts might help to guide ChatGPT's generation and further improve its performance and will investigate it in the future. 

Moreover, there are examples that ChatGPT demonstrates limited comprehension of given prompts. Fig \ref{fig:rating failure} shows a case of the SummEval dataset in the consistency rating task. Though the summary is short, the fact within it is consistent with the article which ChatGPT also admits in the answer. Therefore, all three experts mark 5 out of 5 for the summary. However, ChatGPT then only rates the summary 1 point as it does not cover other facts in the article which is not in the given marking rubric, showing an inadequate understanding of giving prompts. This example demonstrates the insufficient alignment brought by our tested prompt. Prompt engineering including human-in-the-loop alignment optimization and few-shot in-context learning might be helpful to better calibrate ChatGPT's output.

\section{Conclusion}
In this paper, we comprehensively investigate the factual inconsistency evaluation ability of ChatGPT in the zero-shot setting with three coarse-grained and fine-grained factual inconsistency detection tasks.
Our experimental results empirically show the great potential of ChatGPT as a good factual inconsistency evaluator, where it outperforms SOTA evaluation metrics on six out of nine datasets.
Although its great potential, ChatGPT is also found to have limitations on evaluation bias, false reasoning, and hallucination, which should be further addressed for its reliable use.
The experiments also show that ChatGPT's performance can be significantly boosted by the chain-of-thought prompt.
Lastly, We analyzed the limitation of the chain-of-thought prompt, which highlights the importance of alignment research in future work.
The study in our paper is just the initial step in exploring the factual inconsistency evaluation ability of ChatGPT, which we hope can provide useful insights for future work in this direction.

\section*{Limitations}
Our study has the following limitations: 1) Due to the cost limitation of using the API of ChatGPT, we only investigated the effectiveness of using zero-shot prompts on three tasks. More effective prompts such as the few-shot prompts can be explored in future work; 2) We only evaluated the performance of ChatGPT on the factual inconsistency evaluation. A thorough comparison of different large language models (LLMs) such as GPT-3.5 and GPT-4 can be studied in future work, to help us figure out the superiors and limitations of different LLMs.

\bibliography{anthology,custom}

\begin{thebibliography}{52}
\expandafter\ifx\csname natexlab\endcsname\relax\def\natexlab#1{#1}\fi

\bibitem[{Bang et~al.(2023)Bang, Cahyawijaya, Lee, Dai, Su, Wilie, Lovenia, Ji,
  Yu, Chung et~al.}]{bang2023multitask}
Yejin Bang, Samuel Cahyawijaya, Nayeon Lee, Wenliang Dai, Dan Su, Bryan Wilie,
  Holy Lovenia, Ziwei Ji, Tiezheng Yu, Willy Chung, et~al. 2023.
\newblock A multitask, multilingual, multimodal evaluation of chatgpt on
  reasoning, hallucination, and interactivity.
\newblock \emph{arXiv preprint arXiv:2302.04023}.

\bibitem[{Brodersen et~al.(2010)Brodersen, Ong, Stephan, and
  Buhmann}]{brodersen2010balanced}
Kay~Henning Brodersen, Cheng~Soon Ong, Klaas~Enno Stephan, and Joachim~M
  Buhmann. 2010.
\newblock The balanced accuracy and its posterior distribution.
\newblock In \emph{2010 20th international conference on pattern recognition},
  pages 3121--3124. IEEE.

\bibitem[{Brown et~al.(2020)Brown, Mann, Ryder, Subbiah, Kaplan, Dhariwal,
  Neelakantan, Shyam, Sastry, Askell et~al.}]{brown2020language}
Tom Brown, Benjamin Mann, Nick Ryder, Melanie Subbiah, Jared~D Kaplan, Prafulla
  Dhariwal, Arvind Neelakantan, Pranav Shyam, Girish Sastry, Amanda Askell,
  et~al. 2020.
\newblock Language models are few-shot learners.
\newblock \emph{Advances in neural information processing systems},
  33:1877--1901.

\bibitem[{Chowdhery et~al.(2022)Chowdhery, Narang, Devlin, Bosma, Mishra,
  Roberts, Barham, Chung, Sutton, Gehrmann et~al.}]{chowdhery2022palm}
Aakanksha Chowdhery, Sharan Narang, Jacob Devlin, Maarten Bosma, Gaurav Mishra,
  Adam Roberts, Paul Barham, Hyung~Won Chung, Charles Sutton, Sebastian
  Gehrmann, et~al. 2022.
\newblock Palm: Scaling language modeling with pathways.
\newblock \emph{arXiv preprint arXiv:2204.02311}.

\bibitem[{Durmus et~al.(2020{\natexlab{a}})Durmus, He, and
  Diab}]{durmus-etal-2020-feqa}
Esin Durmus, He~He, and Mona Diab. 2020{\natexlab{a}}.
\newblock \href {https://doi.org/10.18653/v1/2020.acl-main.454} {{FEQA}: A
  question answering evaluation framework for faithfulness assessment in
  abstractive summarization}.
\newblock In \emph{Proceedings of the 58th Annual Meeting of the Association
  for Computational Linguistics}, pages 5055--5070, Online. Association for
  Computational Linguistics.

\bibitem[{Durmus et~al.(2020{\natexlab{b}})Durmus, He, and Diab}]{Durmus2020}
Esin Durmus, He~He, and Mona~T Diab. 2020{\natexlab{b}}.
\newblock Feqa: A question answering evaluation framework for faithfulness
  assessment in abstractive summarization.
\newblock \emph{ArXiv}, abs/2005.03754.

\bibitem[{Fabbri et~al.(2021)Fabbri, Kryściński, McCann, Xiong, Socher, and
  Radev}]{Fabbri2021}
Alexander~R Fabbri, Wojciech Kryściński, Bryan McCann, Caiming Xiong, Richard
  Socher, and Dragomir Radev. 2021.
\newblock Summeval: Re-evaluating summarization evaluation.
\newblock \emph{Transactions of the Association for Computational Linguistics},
  9:391--409.

\bibitem[{Falke et~al.(2019{\natexlab{a}})Falke, Ribeiro, Utama, Dagan, and
  Gurevych}]{falke-etal-2019-ranking}
Tobias Falke, Leonardo F.~R. Ribeiro, Prasetya~Ajie Utama, Ido Dagan, and Iryna
  Gurevych. 2019{\natexlab{a}}.
\newblock \href {https://doi.org/10.18653/v1/P19-1213} {Ranking generated
  summaries by correctness: An interesting but challenging application for
  natural language inference}.
\newblock In \emph{Proceedings of the 57th Annual Meeting of the Association
  for Computational Linguistics}, pages 2214--2220, Florence, Italy.
  Association for Computational Linguistics.

\bibitem[{Falke et~al.(2019{\natexlab{b}})Falke, Ribeiro, Utama, Dagan, and
  Gurevych}]{Falke2019}
Tobias Falke, Leonardo F~R Ribeiro, Prasetya~Ajie Utama, Ido Dagan, and Iryna
  Gurevych. 2019{\natexlab{b}}.
\newblock Ranking generated summaries by correctness: An interesting but
  challenging application for natural language inference.

\bibitem[{Fu et~al.(2023)Fu, Ng, Jiang, and Liu}]{Fu2023GPTScoreEA}
Jinlan Fu, See-Kiong Ng, Zhengbao Jiang, and Pengfei Liu. 2023.
\newblock Gptscore: Evaluate as you desire.
\newblock \emph{ArXiv}, abs/2302.04166.

\bibitem[{Goodrich et~al.(2019)Goodrich, Rao, Liu, and
  Saleh}]{goodrich2019assessing}
Ben Goodrich, Vinay Rao, Peter~J Liu, and Mohammad Saleh. 2019.
\newblock Assessing the factual accuracy of generated text.
\newblock In \emph{Proceedings of the 25th ACM SIGKDD International Conference
  on Knowledge Discovery \& Data Mining}, pages 166--175.

\bibitem[{Goyal and Durrett(2020)}]{Goyal2020}
Tanya Goyal and Greg Durrett. 2020.
\newblock \href {https://doi.org/10.18653/v1/2020.findings-emnlp.322}
  {Evaluating factuality in generation with dependency-level entailment}.
\newblock pages 3592--3603. Association for Computational Linguistics (ACL).

\bibitem[{Huang et~al.(2020)Huang, Cui, Yang, Bao, Wang, Xie, and
  Zhang}]{huang-etal-2020-achieved}
Dandan Huang, Leyang Cui, Sen Yang, Guangsheng Bao, Kun Wang, Jun Xie, and Yue
  Zhang. 2020.
\newblock \href {https://doi.org/10.18653/v1/2020.emnlp-main.33} {What have we
  achieved on text summarization?}
\newblock In \emph{Proceedings of the 2020 Conference on Empirical Methods in
  Natural Language Processing (EMNLP)}, pages 446--469, Online. Association for
  Computational Linguistics.

\bibitem[{Huang et~al.(2021)Huang, Feng, Feng, and Qin}]{huang2021factual}
Yichong Huang, Xiachong Feng, Xiaocheng Feng, and Bing Qin. 2021.
\newblock The factual inconsistency problem in abstractive text summarization:
  A survey.
\newblock \emph{arXiv preprint arXiv:2104.14839}.

\bibitem[{Jiao et~al.(2023)Jiao, Wang, Huang, Wang, and Tu}]{jiao2023chatgpt}
Wenxiang Jiao, Wenxuan Wang, Jen-tse Huang, Xing Wang, and Zhaopeng Tu. 2023.
\newblock Is chatgpt a good translator? a preliminary study.
\newblock \emph{arXiv preprint arXiv:2301.08745}.

\bibitem[{Kendall(1938)}]{kendall1938new}
Maurice~G Kendall. 1938.
\newblock A new measure of rank correlation.
\newblock \emph{Biometrika}, 30(1/2):81--93.

\bibitem[{Kocmi and Federmann(2023{\natexlab{a}})}]{Kocmi2023LargeLM}
Tom Kocmi and Christian Federmann. 2023{\natexlab{a}}.
\newblock Large language models are state-of-the-art evaluators of translation
  quality.
\newblock \emph{ArXiv}, abs/2302.14520.

\bibitem[{Kocmi and Federmann(2023{\natexlab{b}})}]{Kocmi2023}
Tom Kocmi and Christian Federmann. 2023{\natexlab{b}}.
\newblock Large language models are state-of-the-art evaluators of translation
  quality.
\newblock \emph{ArXiv}, abs/2302.14520.

\bibitem[{Kojima et~al.(2022)Kojima, Gu, Reid, Matsuo, and
  Iwasawa}]{Kojima2022}
Takeshi Kojima, Shixiang~Shane Gu, Machel Reid, Yutaka Matsuo, and Yusuke
  Iwasawa. 2022.
\newblock Large language models are zero-shot reasoners.
\newblock \emph{ArXiv}, abs/2205.11916.

\bibitem[{Kryściński et~al.(2020)Kryściński, McCann, Xiong, and
  Socher}]{Wojciech2020}
Wojciech Kryściński, Bryan McCann, Caiming Xiong, and Richard Socher. 2020.
\newblock \href {https://doi.org/10.18653/v1/2020.emnlp-main.750} {Evaluating
  the factual consistency of abstractive text summarization}.
\newblock pages 9332--9346. Association for Computational Linguistics (ACL).

\bibitem[{Laban et~al.(2021)Laban, Schnabel, Bennett, and
  Hearst}]{laban-etal-2021-keep}
Philippe Laban, Tobias Schnabel, Paul Bennett, and Marti~A. Hearst. 2021.
\newblock \href {https://doi.org/10.18653/v1/2021.acl-long.498} {Keep it
  simple: Unsupervised simplification of multi-paragraph text}.
\newblock In \emph{Proceedings of the 59th Annual Meeting of the Association
  for Computational Linguistics and the 11th International Joint Conference on
  Natural Language Processing (Volume 1: Long Papers)}, pages 6365--6378,
  Online. Association for Computational Linguistics.

\bibitem[{Laban et~al.(2022)Laban, Schnabel, Bennett, and
  Hearst}]{laban2022summac}
Philippe Laban, Tobias Schnabel, Paul~N Bennett, and Marti~A Hearst. 2022.
\newblock Summac: Re-visiting nli-based models for inconsistency detection in
  summarization.
\newblock \emph{Transactions of the Association for Computational Linguistics},
  10:163--177.

\bibitem[{Lewis et~al.(2020)Lewis, Liu, Goyal, Ghazvininejad, Mohamed, Levy,
  Stoyanov, and Zettlemoyer}]{lewis-etal-2020-bart}
Mike Lewis, Yinhan Liu, Naman Goyal, Marjan Ghazvininejad, Abdelrahman Mohamed,
  Omer Levy, Veselin Stoyanov, and Luke Zettlemoyer. 2020.
\newblock \href {https://doi.org/10.18653/v1/2020.acl-main.703} {{BART}:
  Denoising sequence-to-sequence pre-training for natural language generation,
  translation, and comprehension}.
\newblock In \emph{Proceedings of the 58th Annual Meeting of the Association
  for Computational Linguistics}, pages 7871--7880, Online. Association for
  Computational Linguistics.

\bibitem[{Lin(2004)}]{lin-2004-rouge}
Chin-Yew Lin. 2004.
\newblock \href {https://aclanthology.org/W04-1013} {{ROUGE}: A package for
  automatic evaluation of summaries}.
\newblock In \emph{Text Summarization Branches Out}, pages 74--81, Barcelona,
  Spain. Association for Computational Linguistics.

\bibitem[{Liu and Lapata(2019)}]{liu-lapata-2019-text}
Yang Liu and Mirella Lapata. 2019.
\newblock \href {https://doi.org/10.18653/v1/D19-1387} {Text summarization with
  pretrained encoders}.
\newblock In \emph{Proceedings of the 2019 Conference on Empirical Methods in
  Natural Language Processing and the 9th International Joint Conference on
  Natural Language Processing (EMNLP-IJCNLP)}, pages 3730--3740, Hong Kong,
  China. Association for Computational Linguistics.

\bibitem[{Liu et~al.(2019)Liu, Ott, Goyal, Du, Joshi, Chen, Levy, Lewis,
  Zettlemoyer, and Stoyanov}]{Liu2019RoBERTaAR}
Yinhan Liu, Myle Ott, Naman Goyal, Jingfei Du, Mandar Joshi, Danqi Chen, Omer
  Levy, Mike Lewis, Luke Zettlemoyer, and Veselin Stoyanov. 2019.
\newblock Roberta: A robustly optimized bert pretraining approach.
\newblock \emph{ArXiv}, abs/1907.11692.

\bibitem[{Maynez et~al.(2020)Maynez, Narayan, Bohnet, and
  McDonald}]{Maynez2020}
Joshua Maynez, Shashi Narayan, Bernd Bohnet, and Ryan~T McDonald. 2020.
\newblock On faithfulness and factuality in abstractive summarization.
\newblock \emph{ArXiv}, abs/2005.00661.

\bibitem[{Mishra et~al.(2021)Mishra, Patel, Vijayakumar, Li, Kapanipathi, and
  Talamadupula}]{mishra-etal-2021-looking}
Anshuman Mishra, Dhruvesh Patel, Aparna Vijayakumar, Xiang~Lorraine Li, Pavan
  Kapanipathi, and Kartik Talamadupula. 2021.
\newblock \href {https://doi.org/10.18653/v1/2021.naacl-main.104} {Looking
  beyond sentence-level natural language inference for question answering and
  text summarization}.
\newblock In \emph{Proceedings of the 2021 Conference of the North American
  Chapter of the Association for Computational Linguistics: Human Language
  Technologies}, pages 1322--1336, Online. Association for Computational
  Linguistics.

\bibitem[{Mukaka(2012)}]{mukaka2012guide}
Mavuto~M Mukaka. 2012.
\newblock A guide to appropriate use of correlation coefficient in medical
  research.
\newblock \emph{Malawi medical journal}, 24(3):69--71.

\bibitem[{Nallapati et~al.(2016)Nallapati, Zhou, Gulcehre, Xiang
  et~al.}]{Nallapati2016}
Ramesh Nallapati, Bowen Zhou, Caglar Gulcehre, Bing Xiang, et~al. 2016.
\newblock Abstractive text summarization using sequence-to-sequence rnns and
  beyond.
\newblock \emph{arXiv preprint arXiv:1602.06023}.

\bibitem[{Nan et~al.(2021)Nan, Nallapati, Wang, Nogueira~dos Santos, Zhu,
  Zhang, McKeown, and Xiang}]{nan-etal-2021-entity}
Feng Nan, Ramesh Nallapati, Zhiguo Wang, Cicero Nogueira~dos Santos, Henghui
  Zhu, Dejiao Zhang, Kathleen McKeown, and Bing Xiang. 2021.
\newblock \href {https://doi.org/10.18653/v1/2021.eacl-main.235} {Entity-level
  factual consistency of abstractive text summarization}.
\newblock In \emph{Proceedings of the 16th Conference of the European Chapter
  of the Association for Computational Linguistics: Main Volume}, pages
  2727--2733, Online. Association for Computational Linguistics.

\bibitem[{OpenAI(2022)}]{ChatGPT2022}
OpenAI. 2022.
\newblock Chatgpt.
\newblock \emph{https://openai.com/blog/chatgpt}.

\bibitem[{Ouyang et~al.(2022)Ouyang, Wu, Jiang, Almeida, Wainwright, Mishkin,
  Zhang, Agarwal, Slama, Ray, Schulman, Hilton, Kelton, Miller, Simens, Askell,
  Welinder, Christiano, Leike, and Lowe}]{Ouyang2022TrainingLM}
Long Ouyang, Jeff Wu, Xu~Jiang, Diogo Almeida, Carroll~L. Wainwright, Pamela
  Mishkin, Chong Zhang, Sandhini Agarwal, Katarina Slama, Alex Ray, John
  Schulman, Jacob Hilton, Fraser Kelton, Luke~E. Miller, Maddie Simens, Amanda
  Askell, Peter Welinder, Paul~Francis Christiano, Jan Leike, and Ryan~J. Lowe.
  2022.
\newblock Training language models to follow instructions with human feedback.
\newblock \emph{ArXiv}, abs/2203.02155.

\bibitem[{Pagnoni et~al.(2021)Pagnoni, Balachandran, and
  Tsvetkov}]{Pagnoni2021}
Artidoro Pagnoni, Vidhisha Balachandran, and Yulia Tsvetkov. 2021.
\newblock \href {https://doi.org/10.18653/v1/2021.naacl-main.383}
  {Understanding factuality in abstractive summarization with frank: A
  benchmark for factuality metrics}.

\bibitem[{Qin et~al.(2023)Qin, Zhang, Zhang, Chen, Yasunaga, and
  Yang}]{qin2023chatgpt}
Chengwei Qin, Aston Zhang, Zhuosheng Zhang, Jiaao Chen, Michihiro Yasunaga, and
  Diyi Yang. 2023.
\newblock Is chatgpt a general-purpose natural language processing task solver?
\newblock \emph{arXiv preprint arXiv:2302.06476}.

\bibitem[{Qin et~al.(2022)Qin, Yuan, Neubig, and Liu}]{qin2022t5score}
Yiwei Qin, Weizhe Yuan, Graham Neubig, and Pengfei Liu. 2022.
\newblock T5score: Discriminative fine-tuning of generative evaluation metrics.
\newblock \emph{arXiv preprint arXiv:2212.05726}.

\bibitem[{Scao et~al.(2022)Scao, Fan, Akiki, Pavlick, and et~al}]{Scao2022}
Teven~Le Scao, Angela Fan, Christopher Akiki, Elizabeth-Jane Pavlick, and
  Suzana~Ili'c et~al. 2022.
\newblock Bloom: A 176b-parameter open-access multilingual language model.
\newblock \emph{ArXiv}, abs/2211.05100.

\bibitem[{Scialom et~al.(2021)Scialom, Dray, Gallinari, Lamprier, Piwowarski,
  Staiano, and Wang}]{Scialom2021}
Thomas Scialom, Paul-Alexis Dray, Patrick Gallinari, Sylvain Lamprier, Benjamin
  Piwowarski, Jacopo Staiano, and Alex Wang. 2021.
\newblock Questeval: Summarization asks for fact-based evaluation.

\bibitem[{Soni and Wade(2023)}]{Soni2023ComparingAS}
Mayank Soni and Vincent~P. Wade. 2023.
\newblock Comparing abstractive summaries generated by chatgpt to real
  summaries through blinded reviewers and text classification algorithms.

\bibitem[{Wang et~al.(2020)Wang, Cho, and Lewis}]{wang-etal-2020-asking}
Alex Wang, Kyunghyun Cho, and Mike Lewis. 2020.
\newblock \href {https://doi.org/10.18653/v1/2020.acl-main.450} {Asking and
  answering questions to evaluate the factual consistency of summaries}.
\newblock In \emph{Proceedings of the 58th Annual Meeting of the Association
  for Computational Linguistics}, pages 5008--5020, Online. Association for
  Computational Linguistics.

\bibitem[{Wang et~al.(2023{\natexlab{a}})Wang, Liang, Meng, Li, Qu, and
  Zhou}]{wang2023cross}
Jiaan Wang, Yunlong Liang, Fandong Meng, Zhixu Li, Jianfeng Qu, and Jie Zhou.
  2023{\natexlab{a}}.
\newblock Cross-lingual summarization via chatgpt.
\newblock \emph{arXiv preprint arXiv:2302.14229}.

\bibitem[{Wang et~al.(2023{\natexlab{b}})Wang, Liang, Meng, Shi, Li, Xu, Qu,
  and Zhou}]{wang2023chatgpt}
Jiaan Wang, Yunlong Liang, Fandong Meng, Haoxiang Shi, Zhixu Li, Jinan Xu,
  Jianfeng Qu, and Jie Zhou. 2023{\natexlab{b}}.
\newblock Is chatgpt a good nlg evaluator? a preliminary study.
\newblock \emph{arXiv preprint arXiv:2303.04048}.

\bibitem[{Wang et~al.(2023{\natexlab{c}})Wang, Liang, Meng, Shi, Li, Xu, Qu,
  and Zhou}]{Wang2023IsCA}
Jiaan Wang, Yunlong Liang, Fandong Meng, Haoxiang Shi, Zhixu Li, Jinan Xu,
  Jianfeng Qu, and Jie Zhou. 2023{\natexlab{c}}.
\newblock Is chatgpt a good nlg evaluator? a preliminary study.
\newblock \emph{ArXiv}, abs/2303.04048.

\bibitem[{Wei et~al.(2022)Wei, Wang, Schuurmans, Bosma, hsin Chi, Le, and
  Zhou}]{Wei2022}
Jason Wei, Xuezhi Wang, Dale Schuurmans, Maarten Bosma, Ed~Huai hsin Chi, Quoc
  Le, and Denny Zhou. 2022.
\newblock Chain of thought prompting elicits reasoning in large language
  models.
\newblock \emph{ArXiv}, abs/2201.11903.

\bibitem[{Williams et~al.(2018)Williams, Nangia, and
  Bowman}]{williams-etal-2018-broad}
Adina Williams, Nikita Nangia, and Samuel Bowman. 2018.
\newblock \href {https://doi.org/10.18653/v1/N18-1101} {A broad-coverage
  challenge corpus for sentence understanding through inference}.
\newblock In \emph{Proceedings of the 2018 Conference of the North {A}merican
  Chapter of the Association for Computational Linguistics: Human Language
  Technologies, Volume 1 (Long Papers)}, pages 1112--1122, New Orleans,
  Louisiana. Association for Computational Linguistics.

\bibitem[{Yang et~al.(2023{\natexlab{a}})Yang, Ji, Zhang, Xie, and
  Ananiadou}]{yang2023evaluations}
Kailai Yang, Shaoxiong Ji, Tianlin Zhang, Qianqian Xie, and Sophia Ananiadou.
  2023{\natexlab{a}}.
\newblock On the evaluations of chatgpt and emotion-enhanced prompting for
  mental health analysis.
\newblock \emph{arXiv preprint arXiv:2304.03347}.

\bibitem[{Yang et~al.(2023{\natexlab{b}})Yang, Li, Zhang, Chen, and
  Cheng}]{yang2023exploring}
Xianjun Yang, Yan Li, Xinlu Zhang, Haifeng Chen, and Wei Cheng.
  2023{\natexlab{b}}.
\newblock Exploring the limits of chatgpt for query or aspect-based text
  summarization.
\newblock \emph{arXiv preprint arXiv:2302.08081}.

\bibitem[{Yuan et~al.(2021)Yuan, Neubig, and Liu}]{yuan2021bartscore}
Weizhe Yuan, Graham Neubig, and Pengfei Liu. 2021.
\newblock Bartscore: Evaluating generated text as text generation.
\newblock \emph{Advances in Neural Information Processing Systems},
  34:27263--27277.

\bibitem[{Zar(2005)}]{zar2005spearman}
Jerrold~H Zar. 2005.
\newblock Spearman rank correlation.
\newblock \emph{Encyclopedia of biostatistics}, 7.

\bibitem[{Zhang et~al.(2020)Zhang, Zhao, Saleh, and Liu}]{zhang2020pegasus}
Jingqing Zhang, Yao Zhao, Mohammad Saleh, and Peter Liu. 2020.
\newblock Pegasus: Pre-training with extracted gap-sentences for abstractive
  summarization.
\newblock In \emph{International Conference on Machine Learning}, pages
  11328--11339. PMLR.

\bibitem[{Zhang et~al.()Zhang, Kishore, Wu, Weinberger, and
  Artzi}]{zhangbertscore}
Tianyi Zhang, Varsha Kishore, Felix Wu, Kilian~Q Weinberger, and Yoav Artzi.
\newblock Bertscore: Evaluating text generation with bert.
\newblock In \emph{International Conference on Learning Representations}.

\bibitem[{Zhong et~al.(2023)Zhong, Ding, Liu, Du, and Tao}]{Zhong2023}
Qihuang Zhong, Liang Ding, Juhua Liu, Bo~Du, and Dacheng Tao. 2023.
\newblock Can chatgpt understand too? a comparative study on chatgpt and
  fine-tuned bert.
\newblock \emph{ArXiv}, abs/2302.10198.

\end{thebibliography}
\bibliographystyle{acl_natbib}



\end{document}